\documentclass{article}

\PassOptionsToPackage{numbers}{natbib}
\usepackage[preprint]{nips_2018}

\usepackage[utf8]{inputenc} 
\usepackage[T1]{fontenc}    
\usepackage{hyperref}       
\usepackage{url}            
\usepackage{booktabs}       
\usepackage{amsfonts}       
\usepackage{nicefrac}       
\usepackage{microtype}      
\usepackage{algorithm2e}    
\usepackage{graphicx}       
\usepackage{subcaption}     
\usepackage[preprint]{nips_2018}

\title{A Maximal Heterogeneity Based Clustering Approach for Obtaining Samples}

\author{
  Megha Mishra \\
    Bachelor of Technology\\
  School of Computing Science \& Engineering\\
  VIT, India\\
  \texttt{megha.mishra2014@vit.ac.in} \\
   \And
   Chandrasekaran Anirudh Bhardwaj \\
	Bachelor of Technology\\   
   School of Computing Science \& Engineering\\
   VIT, India\\
  \texttt{canirudh.bhardwaj2014@vit.ac.in} \\     
   \And
   Kalyani Desikan\\
   Professor \\
   School of Advanced Sciences\\
   VIT, India\\
  \texttt{kalyanidesikan@vit.ac.in} \\  
}

\begin{document}

\maketitle

\begin{abstract}
	
Medical and social sciences demand sampling techniques which are robust, reliable, replicable and have the least dissimilarity between the samples obtained. Majority of the applications of sampling use randomized sampling, albeit with stratification where applicable. The randomized technique is not consistent, and may provide different samples each time, and the different samples themselves may not be similar to each other. In this paper, we introduce a novel non-statistical no-replacement sampling technique called Wobbly Center Algorithm, which relies on building clusters iteratively based on maximizing the heterogeneity inside each cluster. The algorithm works on the principle of stepwise building of clusters by finding the points with the maximal distance from the cluster center. The obtained results are validated statistically using Analysis of Variance tests by comparing the samples obtained to check if they are representative of each other. The obtained results generated from running the Wobbly Center algorithm on benchmark datasets when compared against other sampling algorithms indicate the superiority of the Wobbly Center Algorithm. 

\end{abstract}

\section{Introduction}

Sampling as a technique has been studied for a long time, with a rich history of research into it. Sampling is a method to derive a subset of data from the original population, such that the subset preserves the characteristics of the entirety of the original population.

Fields such as Medical \citenum{Ref 2} and Social Sciences require robust no-replacement sampling techniques to ensure the validity of their hypothesis testing. This means that the samples derived from the original population must have a high degree of variance encapsulated within them to capture the entire characteristics of the original population, but also at the same time have enough similarity between the different samples generated. Further, the results must be replicable.

Random Sampling  \citenum{Ref 2} \citenum{Ref 3}\citenum{Ref 4}\citenum{Ref 5}  is one of the most popular approach to derive samples from the given data. Inherently it is fully random in nature, though replicability can be induced in practical aspect by setting the seed of the pseudo-random number generator used to generate the samples. 

Typically, clustering algorithms \citenum{Ref 6}  have been used to segment the population into partitions which have the most similarity in the points contained in them by maximizing the homogeneity of points inside each cluster. The proposed Wobbly Center Algorithm uses a contrarian approach to clustering, by maximizing the heterogeneity inside each cluster to ensure that each cluster would be able to capture the entire variance of the original population yielding clusters which are similar to each other in excess to the original population itself. This approach is inherently replicable, and this work showcases this approach and validates it statistically over well known benchmark datasets.

\section{Wobbly Center Algorithm}

Wobbly Center Algorithm works on the principle of itereatively building clusters on the basis of maximizing the dissimilarity inside each cluster. Z-Score standardization is used as a pre-processing step to ensure proper scaling and normality of the features, as the proposed approach is scale variant.

Let \\
$\{S\}$ be the set of all the data points in the original population,\\ 
$k$ be the number of samples needed, \\
$\{V\}$ be the set of all data points that have been already assigned to any cluster. \\
$\{C_{1}\}$,$\{C_{2}\}$,..$\{C_{k}\}$ be the set of all samples (Clusters in this specific case) \\
$|\{W\}|$ denote the number of elements in the set $\{W\}$ \\
$|B-A|$ denote the Euclidean Distance of vector B from vector A in the spatial space\\
$\{A\}-\{B\}$ denote the subtraction of set B from set A \\
$\{A\}+\{B\}$ denote the addition of set B with set A \\

\begin{algorithm}[H]
 \KwData{$\{S\}$, $k$}
 \KwResult{$\{C_{1}\}$,$\{C_{2}\}$,..$\{C_{k}\}$}
 Initialization\;
 $\{C_{1}\}$,$\{C_{2}\}$,..$\{C_{k}\}$ <- $\emptyset$ \;
 $\{V\}$ <- $\emptyset $ \;
Algorithm\;
 Find $M$ such that $min_{\forall X\in\{S\}}(|X - M|)$ \; 
 $i$ <- 1\;
\While{$i \leq k$}{
	$X_{min}$ <- Find X such that $min_{\forall X\in\{S\}}(|X-M|)$ \;
	$\{C_i\}$ <- $\{C_i\} + \{X_{min}\} $ \;
	$\{S\}$ <- $\{S\} - \{X_{min}\} $ \;
	$\{V\}$ <- $\{V\} + \{X_{min}\} $ \;
	$i$ <- $i +1$ \;
} 

\While{$ \{S\} \neq \emptyset$}{
	$i$ <- 1 \;
	\While{$i \leq k$}{
		$length$ <- $ | \{C_i\} |$ \;
		$M_i$ <- $\frac{1}{length} \sum_{Y \in \{C_i\}} Y$ \;
		$X_{max}$ <- Find X such that $max_{\forall X\in\{S\}}(|X-M_i|)$ \;
		$\{C_i\}$ <- $\{C_i\} + \{X_{max}\} $ \;
		$\{S\}$ <- $\{S\} - \{X_{max}\} $ \;
		$\{V\}$ <- $\{V\} + \{X_{max}\} $ \;
		$i$ <- $i +1$ \;
	
	}
}
 
return ($\{C_{1}\}$,$\{C_{2}\}$,..$\{C_{k}\}$)
 \caption{Wobbly Center Algorithm}
\end{algorithm}

The algorithm can be roughly split into three main components:
\begin{enumerate}
	\item Seed selection: The points nearest to the mean vector of the entire population are chosen as the seed points
	\item Cluster assignment: The datapoints furtherst away from the cluster mean vector are added to the cluster, and the mean vector is recomputed after each iteration
	\item Termination: The algorithm stops when there are no points left to be added to any one of the clusters
\end{enumerate}

Algorithm 1. describes the working of Wobbly Center Algorithm in detail. The seed points are chosen as close to the center of the scaled dataset as possible to ensure faster convergence. The algorithm stops when there are no additional datapoints in the set $\{S\}$.

As the algorithm progresses over iterations, the mean and variance of the individual clusters begins to resemble the distribution of the original population, eventually converging to the exact value of the original population. This can be seen in Figure 1(a) and Figure 1(b), where the means of the two clusters begins to converge within a few iterations. 

\begin{figure}[h!]
  \centering
  \begin{subfigure}[b]{0.45\linewidth}
    \includegraphics[width=6.5cm]{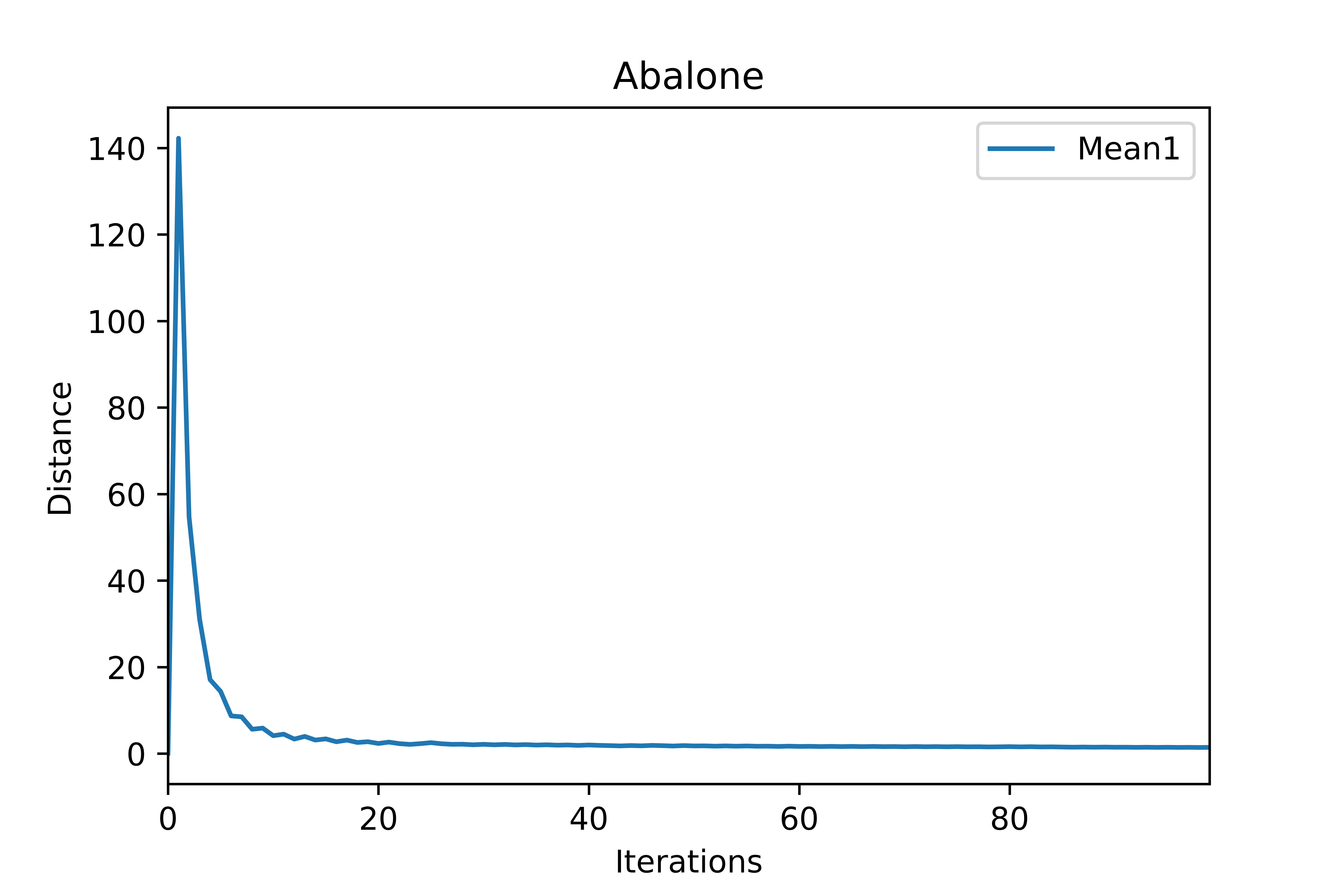}
    \caption{Cluster 1}
  \end{subfigure}
  \begin{subfigure}[b]{0.45\linewidth}
    \includegraphics[width=6.5cm]{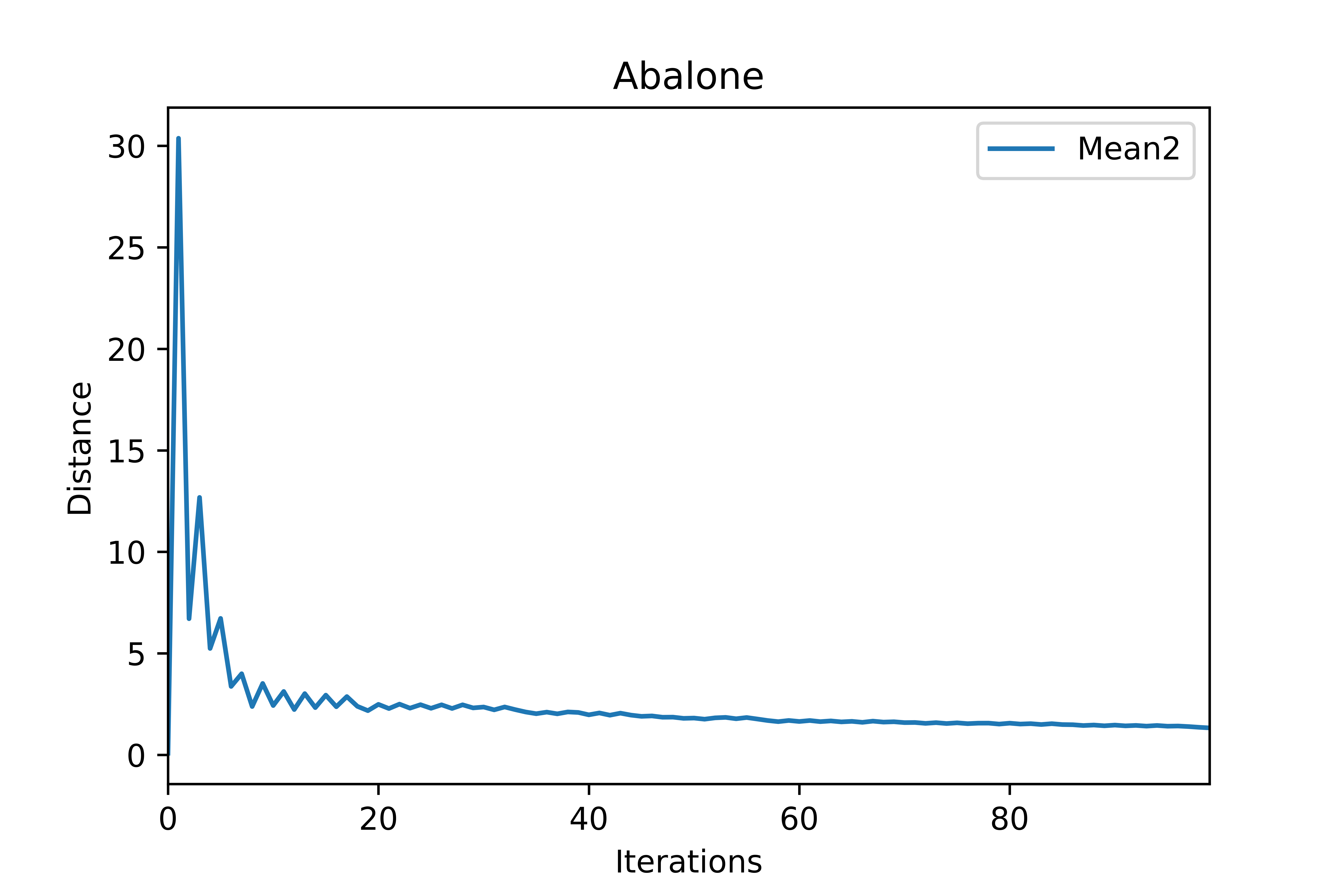}
    \caption{Cluster 2}
  \end{subfigure}
  \caption{Convergence of clusters in Wobbly Center Algorithm for Abalone Dataset\citenum{Ref 15}}
  \label{fig:coffee}
\end{figure}

\section{Experimental Analysis}

\subsection{Test Conditions}
The data was scaled using Z-score standardization, transforming the features to a Normal distribution with 
$$Zscore = \frac{(X-\mu)}{\sigma}$$
where $\mu$ is the mean and $\sigma$ is the standard deviation of the feature. Performing Z-score standardization \citenum{Ref 7} over all the features respectively yields a dataset with features which roughly belong to the Normal distribution with 0 mean and unit variance.
$$Features \sim N(0,1)$$
This scaling is done to ensure the assumptions for the statistical tests are valid, furthermore it is also required as the Wobbly Center Algorithm is scale sensitive.

The datasets were procured from the repository hosted by University of California, Irvine\citenum{Ref 8}.
The random sampling was performed without replacement. The Random Sampling  \citenum{Ref 2} \citenum{Ref 3}\citenum{Ref 4}\citenum{Ref 5} and Z-score Standardization\citenum{Ref 7} were performed using Sci-kit library\citenum{Ref 9} in Python. 
One way Analysis of Variance (ANOVA) \citenum{Ref 10} \citenum{Ref 11} tests used for testing hypothesis were performed using SciPy library \citenum{Ref 12}in Python. 

Two samples of roughly the same sample sizes, corresponding to half the size of the original population were derived using the Wobbly Center Algorithm and Random Sampling based approaches respectively.

\subsection{Results}

Setting the null hypothesis \citenum{Ref 13} as:

$Null ~Hypothesis ~1: ~There ~is ~no ~statistical ~difference ~between ~the ~samples ~and ~the ~original ~population$

$Null ~Hypothesis ~2: ~There ~is ~no ~statistical ~difference ~between ~the ~samples ~obtained$

Further, setting the confidence interval as 95\%, meaning 
$$\alpha=0.05$$
and we reject the null hypothesis if 
$$
p < \alpha $$
In this case,
$$
p < 0.05
$$

Intuitively, $Null ~Hypothesis ~1$ can be proven for both Random Sampling method and Wobbly Center Algorithm given a sufficiently large sample size.

Analysis of Variance (ANOVA) \citenum{Ref 10}\citenum{Ref 11} test is used to analyze the samples with each other to determine if they are statistically similar, in effect checking if $Null ~Hypotheisis ~2$ is valid. In the case where the $k=2$, ANOVA is fundamentally equivalent to Student's t-test \citenum{Ref 14}. One-way ANOVA test is sufficient for proving the similarity of the samples as the interaction between the samples is not relevant.

\begin{table}
  \caption{One-way ANOVA Value for Abalone Dataset\citenum{Ref 15}}
  \label{ANOVA-Result}
  \centering
  \begin{tabular}{lll}
    \toprule
    Attribute     & Random Sampling & Wobbly Center Algorithm  \\
    \midrule
    1	& 0.0239813 & 0.9939368  \\
    2   & 0.0184019 & 0.9947614  \\
    3   & 0.01586   & 0.9703539  \\
    4 	& 0.006387  & 0.983332   \\
    5   & 0.0063109 & 0.9586654  \\
    6   & 0.0097414 & 0.995823   \\
    7 	& 0.0045467 & 0.9886798  \\
    8   & 0.0250208 & 0.9734603  \\
    
    \bottomrule
  \end{tabular}
\end{table}

Table 1. indicates the effectiveness of the Wobbly Center Algorithm over Random Sampling based methods by comparing the p-values obtained by performing the ANOVA test \citenum{Ref 10}\citenum{Ref 11}  on each feature. The $Null ~Hypothesis ~2$ stands rejected for multiple features taken from samples generated by the Random Sampling based approach\citenum{Ref 2} \citenum{Ref 3}\citenum{Ref 4}\citenum{Ref 5}, while it is not possible to reject $Null ~Hypothesis ~2$ for any one of the features taken from samples generated by the Wobbly Center Algorithm. Similar results are observed in other benchmark datasets, an example of which can be seen in Table 2 which covers the Wine dataset\citenum{Ref 16}.

\begin{table}
  \caption{One-way ANOVA Value for Wine Dataset \citenum{Ref 16}}
  \label{ANOVA-Result}
  \centering
  \begin{tabular}{lll}
  \toprule
    Attribute     & Random Sampling & Wobbly Center Algorithm  \\
    \midrule
    1	& 0.92648595 	& 0.93455995  \\
    2   & 0.04170006 	& 0.93053714  \\
    3   & 0.30828398   	& 0.97269005  \\
    4 	& 0.81351111  	& 0.97719202  \\
    5   & 0.30673158 	& 0.97339625  \\
    6   & 0.85206353 	& 0.9644888   \\
    7 	& 0.6971902 	& 0.99664723  \\
    8   & 0.52184036 	& 0.96136958  \\
    9   & 0.1018266 	& 0.94343048  \\
    10  & 0.40231485 	& 0.96106212  \\
    11  & 0.98721795 	& 0.94847321  \\
    
    \bottomrule
  \end{tabular}
\end{table}

\section{Discussion and Conclusion}

The superiority of Wobbly Center Algorithm over the conventional Random Sampling method \citenum{Ref 2} \citenum{Ref 3}\citenum{Ref 4}\citenum{Ref 5}was demonstrated using results derived from conducting one-way ANOVA tests between the samples generated. Intuitively, the Wobbly Center Algorithm can be understood as greedily adding points to clusters on the basis of maximal heterogeneity. 

The Wobbly Center Algorithm could be of great usage in the sphere of medical\citenum{Ref 1} and social sciences, where there is a distinct need to ensure high similarity between the control and experimental groups, to rule out any external influencing factors. Further research is needed to check if hierarchical splitting of clusters would be more optimal than simply splitting the clusters at the same time when conducting the experiment for $k$ > 2.

\end{document}